\documentclass[runningheads]{llncs}
\usepackage{graphicx}
\usepackage[utf8]{inputenc} % allow utf-8 input
\usepackage[T1]{fontenc}    % use 8-bit T1 fonts
\usepackage{hyperref}       % hyperlinks
\usepackage{url}            % simple URL typesetting
\usepackage{booktabs}       % professional-quality tables
\usepackage{amsfonts}       % blackboard math symbols
\usepackage{nicefrac}       % compact symbols for 1/2, etc.
\usepackage{microtype}      % microtypography
\usepackage{csvsimple}
\usepackage{booktabs} % For \toprule, \midrule and \bottomrule
\usepackage{siunitx} % Formats the units and values
\usepackage[table,xcdraw]{xcolor}
\usepackage{makecell}

\begin{document}
\title{Vector-Based Data Improves Left-Right Eye-Tracking Classifier Performance After a Covariate Distributional Shift}
\titlerunning{Improved Eye-Tracking Classifier Performance}
% If the paper title is too long for the running head, you can set
% an abbreviated paper title here
%
\author{Brian Xiang\inst{1}\orcidID{0000-0002-3403-2014}
\and
Abdelrahman Abdelmonsef\inst{1}\orcidID{
0000-0001-6952-2681}
\thanks{The two authors contributed equally to the paper}}
\authorrunning{Xiang and Abdelmonsef}
% First names are abbreviated in the running head.
% If there are more than two authors, 'et al.' is used.
%
\institute{Swarthmore College, Swarthmore PA 19081, USA\\
\email{\{bxiang1,ayahia1\}@swarthmore.edu}   }
\maketitle              % typeset the header of the contribution
\begin{abstract}
The main challenges of using electroencephalogram (EEG) signals to make eye-tracking (ET) predictions are the differences in distributional patterns between benchmark data and real-world data and the noise resulting from the unintended interference of brain signals from multiple sources. Increasing the robustness of machine learning models in predicting eye-tracking position from EEG data is therefore integral for both research and consumer use. In medical research, the usage of more complicated data collection methods to test for simpler tasks has been explored to address this very issue. In this study, we propose a fine-grain data approach for EEG-ET data collection in order to create more robust benchmarking. We train machine learning models utilizing both coarse-grain and fine-grain data and compare their accuracies when tested on data of similar/different distributional patterns in order to determine how susceptible EEG-ET benchmarks are to differences in distributional data. We apply a covariate distributional shift to test for this susceptibility. Results showed that models trained on fine-grain, vector-based data were less susceptible to distributional shifts than models trained on coarse-grain, binary-classified data.


\end{abstract}
\section{Introduction}
\subsection{Problem Statement}
Eye-tracking (ET) is the process of predicting a subject’s point of gaze or robustly detecting the relative motion of the eye to the head \cite{klaib2021eye}. Electroencephalography (EEG) signals are brain signals that correspond to various states from the scalp surface area, encoding neurophysiological markers, and are characterized by their high temporal resolution and minimal restrictions \cite{kumar2012analysis}.

Recently, predicting eye-tracking data using EEG has received increasing interest because of the recent hardware and software technological advancements and its versatile applications in different fields \cite{sabanci2015classification,hollenstein2018zuco,kastrati2021eegeyenet}. For instance, EEG-ET data can be used to identify shopping motives relatively early in the search process \cite{pfeiffer2020eye}, to detect workload strain in truck drivers \cite{lobo2016cognitive}, and to assess the diagnosis of many neurological diseases such as Autism Spectrum Disorder and Alzheimer's \cite{thapaliya2018evaluating,sotoodeh2021preserved,kang2020identification}. 

With the broad array of EEG-ET data applications, accuracy and efficiency are of utmost importance. The collection of both EEG and ET data, however, involves challenges such as finding invariant representation of inter-and intra-subject differences as well as noise resulting from the unintended interference of brain signals from multiple sources \cite{wu2022learning}. Increasing the robustness of machine learning models in making accurate eye-tracking predictions from EEG data is therefore integral for both research and consumer use.

\subsection{Literature Review}
EEG is a type of highly individualized time-series data \cite{qu2018personalized,qu2018eeg,qu2020identifying,roy2019deep}. Machine Learning approaches perform adequately on computer vision, bioinfomatics, medical image analysis and usually have potential in EEG analysis as well \cite{qian2021two,qian2020multi,gu2020multi,xu2020multi}. Machine Learning and Deep Learning methods have been implemented in EEG classification research with promising results \cite{lotte2018review,qu2020multi,qu2020using,craik2019deep}. Recent works in this area have focused on determining what machine learning models are more capable of predicting eye position from EEG signals. In that pursuit, \cite{lotte2018review} has shown that Riemannian geometry and tensor-based classification methods have reached state-of-the-art performances in multiple EEG-based machine learning applications. Previous work has also been done in collecting large datasets that combine EEG and ET data. A good example is the multi-modal neurophysiological dataset collected by \cite{langer2017resource} and EEGEyeNet \cite{kastrati2021eegeyenet}. Finally, other work has been done to eliminate the noise associated with EEG data collection automatically, such as \cite{plochl2012combining,oikonomou2020machine}. Also, \cite{roy2019machine} showed the potential of using machine learning for noise elimination. Soon after, Zhang et al. collected EEGdenoiseNet, a dataset suitable to train machine learning models in noise elimination \cite{zhang2021eegdenoisenet}. These benchmarks include ocular artifacts; thus, they apply to EEG data collected simultaneously with eye-tracking data. 

While much work has been done to improve the accuracy of ET-based EEG applications, there is a lack of previous work that focuses on examining the effects of data collection methods on eye position prediction accuracies. In particular, we are interested in comparing coarse-grain and fine-grain data collection techniques for simultaneously collecting EEG and eye-tracking data. In other fields of medical research, fine-grain data collection methods have been explored with great success \cite{higginson2000ecological,plancher2012using}. That is using more complicated data collection methods to test for less complicated tasks. Often, systems developed in "lab conditions" only work in controlled environments, causing difficulties when utilized in uncontrolled conditions \cite{wilson1993,marcotte2010neuropsychology}. With consumer EEG-ET applications becoming increasingly prevalent, the importance of assessing whether finer-grain data collection methods improve upon the accuracy and robustness of EEG-ET machine learning classifiers is magnified. To narrow down this assessment, this study focuses on the effects of training machine learning models on fine- versus coarse-grain EEG data in predicting the binary direction (left or right) of eye-tracking position. 

\subsection{Purpose of Study}
Previously, benchmarks for left-right eye-tracking predictions were established utilizing data from binary-classified data collection methods \cite{kastrati2021eegeyenet}. In this study, we train machine learning models for left-right eye-tracking classification using data from a vector-based collection framework. We then compare the results to the models trained by Kastrati et al. and determine whether finer-grain data is beneficial to the robustness of EEG-ET machine learning models. Robustness is determined by the accuracy after a covariate distributional shift. This attempts to mimic realistic data with varying distributional patterns.
The purpose of this study is to verify whether machine learning models trained using vector-based, fine-grain eye-tracking data obtain higher accuracies in left-right gaze classification than models trained using binary-based, coarse-grain eye-tracking data after a covariate shift is applied. 

\section{Data}
In our experiments, we used the EEGEyeNet dataset  \cite{kastrati2021eegeyenet}. The EEGEyeNet dataset gathers EEG and eye-tracking data from 356 (190 female and 166 male) healthy subjects of varying ages (18 - 80). Data were collected for 3 different experimental tasks according to the Declaration of Helsinki Principles. The tasks are Pro- and Antisaccade, Large Grid, and Visual Symbol Search. The recording setup was similar across all experimental tasks. EEG data were recorded using a 128-channel EEG Geodesic Hydrocel system at a sampling rate of 500 Hz. Eye position was simultaneously recorded using an infrared video-based ET EyeLink 1000 Plus also at a sampling rate of 500 Hz. Between trials, the eye-tracking device would be reset with a 9-point grid. Subjects were situated 68 cm away from a 24-inch monitor with a resolution of 800 x 600 pixels. In this study, we directly use data from the pro-antisaccade paradigm and manipulate data from the large grid paradigm to use for the analysis. 

\subsection{Pro-Antisaccade Task}
\begin{figure} [!t]
    \centering
    \includegraphics[scale = 0.8]{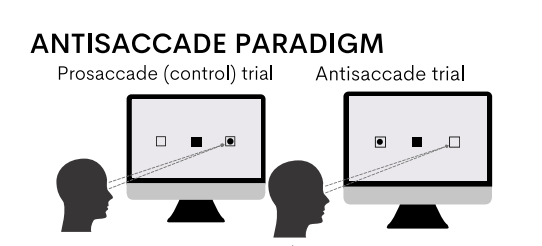}
    \caption{Schematic for the location of the cues on the screen in the pro-antisaccade paradigm \cite{kastrati2021eegeyenet}}
    \label{fig:PAED}
\end{figure}

In this task, each trial starts with a central square on which the participants were asked to focus for a randomized period that doesn't exceed 3.5 seconds. Then, a dot appears horizontally to the left or the right of this central square, as shown in Figure \ref{fig:PAED}. Subjects were asked to perform a saccade towards the opposite side of the cue and then look back to the center of the screen after 1 second. Thus, gaze positions in pro-antisaccade were restricted to the horizontal axis and were binary-classified either left or right (relative to the dot at the screen's center).

\subsection{Large Grid Task}
\begin{figure} [!b]
    \centering
    \includegraphics[scale = 0.8]{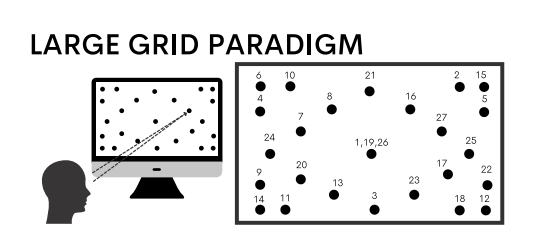}
   \caption{Schematic for the location of the dots on the screen in the large grid paradigm \cite{kastrati2021eegeyenet}}
    \label{fig:LGPAED}
\end{figure}

For this task, subjects were asked to stare at a blank screen and focus on the appearing dots. A set of well-distributed dots in 25 positions would then appear one at a time for 1.5 to 1.8 seconds, as shown in Figure \ref{fig:LGPAED}. The center dot would appear three times for a total of 27 trials per block. Subjects perform 5 blocks and repeat the procedure 6 times for a total of 810 trials per subject. Thus, large grid's framework enabled fine-grain EEG-ET data collection as gaze positions were encoded using both angle and amplitude.

\section{Experiment Design}
The learning objective of the machine learning models trained in our experiment was to use EEG brain signals to predict the direction of a subject's gaze along the horizontal axis (whether they are looking to the left or the right). Although predictions for this task using the same dataset were made by \cite{kastrati2021eegeyenet}, they were performed exclusively using data from the prosaccade trials of the pro-antisaccade task for both training and testing. 

In this paper, we train 8 machine learning models on both the pro-antisaccade and large grid data and perform a covariate distributional shift by comparing their performance based on the accuracy when tested on data from a different experimental task. As per the author's recommendation, we used the minimally preprocessed EEG data. To split the data, data from 70\% of the participants were assigned to the training set, data from 15\% of the participants was assigned to the cross-validation set, and 15\% for the testing. No single participant's data was shared among different sets; each participant's data is entirely assigned to one set. 

\subsection{Data Processing}
Given the classification nature of our learning problem, data from the large grid paradigm, encoded as Angle and Amplitude, should be translated and labeled into the expected format for training left-right classification models. The expected format for our labels, or data in the algorithm we used, was a lookup table with the subject ID to identify unique participants and 1 or 0 to encode left or right. 

\subsubsection{Angle to Direction:}

\begin{figure}[!t]
    \centering
    \includegraphics[scale = 0.3]{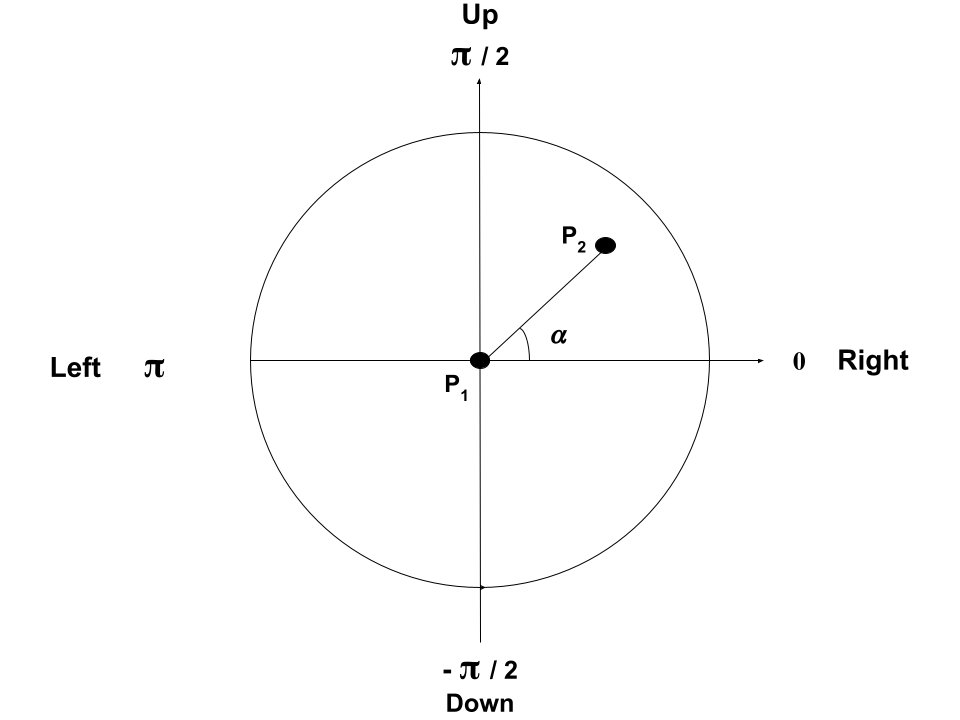}
    \caption{Illustration of angle $\alpha$. $P_1$ represents the initial gazing position of the eye and $P_2$ represents the end gazing position of the eye. The line between them represents the movement of the eye.}
    \label{fig:alpha}
\end{figure}

The angle value for each training set will be used as an indication of the gaze direction. Given that the angle values in the dataset range from $-\pi$ to $\pi$, the translation logic is only concerned about the values in this range; angles with absolute values larger than $\pi$ are ignored because they are not represented in the dataset. To perform the transformation from angle to left and right, the adopted convention for conversion is that for angle $\alpha$, $|\alpha|\leq\pi/2$ will be assigned the right direction; $|\alpha|>\pi/2$ will be assigned the left direction. As shown in Appendix A, the logic for this transformation has been confirmed by the dataset authors and is shown in Figure \ref{fig:alpha}.

\subsubsection{Amplitude and ID:}

Given the role of ID in properly splitting the data, the subject ID will be kept and used the same as in the original format. We plan on adding amplitude later to the translated left-right values to determine if weighting data points increases or decreases accuracy. 

The processing code implementing the logic above can be found \href{https://github.com/brianxiang123/EEGETCovariateDistributionalShift}{here}.

\subsection{Models Training}
Experiments were conducted for 4 different combinations of training and testing datasets: models trained on pro-antisaccade data and tested on pro-antisaccade data, models trained on pro-antisaccade data and tested on large grid data, models trained on large grid data and tested on large grid data, and models trained on large grid data and tested on pro-antisaccade data. The EEGEyeNet code was used as an interface for feature extraction and running the machine learning models on the extracted features. Since our task is a classification problem, regression machine learning models were excluded from running this task. The included models are the Decision Tree, Random Forest, XGBoost, Radial Basis Function kernel SVC, Gradient Boost, Gaussian Naive Bayes, AdaBoost, and Linear Support Vector Classification (SVC). Data processing was done using the NumPy library and the model implementations were installed from the SKlearn library \cite{pedregosa2011scikit}. Specifications regarding the test environment, actual runtime of the models, and total space occupied by the dataset are provided in Appendix B.

\section{Models}
\subsection{Machine Learning Models}
In this study, machine learning models operate on features extracted from the data rather than the data itself. Feature extraction has been applied in two steps. First, \cite{kastrati2021eegeyenet} applied a band-pass filter in frequencies in the range [8 - 13 HZ] on the acquired signals through all trials. This choice of frequencies is based on \cite{foster2017alpha} suggestions. Following the filtering step, the Hilbert transform was applied, resulting in a complex time series from which targeted features were extracted for learning models. Since we are considering a classification problem, we experimented with classification-only models (Linear SVC) and models that can be applied to both classification and regression problems, such as ensemble classifiers.     

\subsubsection{SVCs:} 
SVC is an abbreviation for Support Vector Classifier. These are machine learning algorithms that are commonly used for supervised classification problems
and sometimes regression problems, which are called Support Vector Regression. The classification decision relies on finding the best hyperplanes in the feature space. These planes are, then, used to 
differentiate the predictions’ classes based on their orientation to the planes. A simple example in a 2d feature space might be a line that distinguishes two groups 
and classifies predictions based on whether the point in the 2d feature space lies to the left or the right of the plane. In our paper, we will be using Linear SVC and RBF. These algorithms still perform well after several decades since they were first implemented in this field \cite{bashivan2016mental,lotte2015signal} 

\subsubsection{Ensemble Classifiers} 

Ensemble (or voting) classifier is a machine learning classification algorithm that trains with different classification models and makes predictions through ensembling their predictions to make a stronger classification. These algorithms have been the gold standard for several EEG-based classification experiments \cite{qu2020multi}. In our study, we used Random Forest, XGBoost, GradientBoost, and AdaBoost. 

\subsubsection{Naive Bayes and Decision Tree Classifiers}

Naive Bayes (NB) Classifier is the statistical Bayesian classifier \cite{duda1973pattern}. It assumes that all variables are mutually correlated and contribute to some decree towards classification. It is based on the Bayesian theorem and is commonly used with high dimensional inputs. On the other hand, a decision tree is not a statistically based one; rather, it is a data mining induction technique that recursively partitions the dataset using a depth-first greedy algorithm till all the data gets classified to a particular class. Both NB and the decision tree are relatively fast and well suited to large data. Furthermore, they can deal with noisy data, which makes them well suited for EEG classification applications \cite{jadhav2016comparative}. 

\subsection{Deep Learning Models}
Deep learning is a subfield of machine learning algorithms in which computational models learn features from hierarchical representations of input data through successive non-linear transformations \cite{roy2019deep}. Deep learning methods, especially Convolutional Neural Network (CNN), performed well in several previous EEG band power (feature) based research \cite{qu2020multi}. Still, these methods have not demonstrated convincing and consistent improvements \cite{lotte2018review}. Given so and their high run time, they are excluded from analysis in this paper.

\section{Results}
We trained the machine learning models using the two datasets (pro-antisaccade and "translated" large grid), tested each of them on data from the two datasets, and compared the results. Thus, we had 4 combinations of training and testing datasets: pro-antisaccade - pro-antisaccade (PA-PA), large grid - large grid (LG-LG), pro-antisaccade - large grid (PA-LG), and large grid - pro-antisaccade (LG-PA). PA-PA is essentially a recreation of the work of \cite{kastrati2021eegeyenet} and LG-LG compares whether training and testing on large grid will obtain higher accuracies than that of PA-PA. 

Comparing PA-LG and LG-PA is the main objective of the paper since it will give insight into whether models trained on fine-grain data (the vector-based eye tracking large grid data) will be more accurate when tested against unfamiliar data from the pro-antisaccade experiment, or vice versa. 

Results regarding accuracies, standard deviation for accuracies, and mean run time for each algorithm in all combinations are provided in Tables \ref{tab:LRA}, \ref{tab:LRLGP}, \ref{tab:LRALGP}, and \ref{tab:LRLGPA}. Some of the standard deviations could not be calculated because some of the models did not converge. A summary table of the results is presented in Table \ref{tab:data}. 

\begin{table}[!t]
    \centering
    \caption{Left-right task trained and tested on the pro-antisaccade data}
    \label{tab:LRA}
    \begin{tabular}{|l|l|l|l|l|}
    \hline
        Model & Accuracy (\%) & Standard Deviation & Mean Run time (s) \\ \hline

        XGBoost  \cite{tiwari2019multiclass} & 97.9 & 1.11E-16 & 0.062 \\ \hline
        GradientBoost  \cite{freund1997decision} & 97.4 & 3.8E-4 & 0.016 \\ \hline
        RandomForest  \cite{edla2018classification} & 96.5 & 5.15E-4 & 0.121 \\ \hline
        AdaBoost  \cite{freund1997decision} & 96.3 & 1.11E-16 & 0.142 \\ \hline
        DecisionTree \cite{aydemir2014decision} & 96.2 & 1.11E-16 & 0.012 \\ \hline
        LinearSVC  \cite{bhuvaneswari2013support} & 92.0 & 1.55E-4 & 0.007 \\ \hline
        RBF SVC  \cite{satapathy2017eeg} & 89.4 & N/A & 1.864 \\ \hline
        GaussianNB  \cite{carrion2021analysis} & 87.7 & 1.11E-16 & 0.012 \\ \hline
    \end{tabular}
\end{table}

\begin{table}[!b]
    \centering
    \caption{Left-right task trained and tested on the large grid data}
    \label{tab:LRLGP}
    \begin{tabular}{|l|l|l|l|l|}
    \hline
        Model & Accuracy (\%) & Standard Deviation & Mean Run time (s) \\ \hline
        
        GradientBoost & 95.6 & 3.6E-4 & 0.013  \\ \hline
        XGBoost & 95.5 & 1.11E-16 & 0.082 \\ \hline
        RandomForest & 94.6 & 1.55E-3 & 0.081  \\ \hline
        AdaBoost & 93.5 & N/A & 0.086  \\ \hline
        DecisionTree  & 92.0 & N/A & 0.009 \\ \hline
        LinearSVC & 92.0 & 2.94E-4 & 0.006  \\ \hline
        RBF SVC & 88.9 & N/A & 1.071  \\ \hline
        GaussianNB & 87.2 & N/A & 0.013  \\ \hline
    \end{tabular}
\end{table}

\begin{table}[!t]
    \centering
    \caption{Left-right task trained on pro-antisaccade and tested on large grid}
    \label{tab:LRALGP}
    \begin{tabular}{|l|l|l|l|l|}
    \hline
        Model & Accuracy (\%) & Standard Deviation & Mean Run time (s) \\ \hline
        
        XGBoost & 93.4 & 1.11E-16 & 0.054 \\ \hline
        GradientBoost & 92.1 & 3.8E-4 & 0.019 \\ \hline
        LinearSVC & 91.4 & 1.55E-4 & 0.005 \\ \hline
        RandomForest & 91.2 & 5.15E-4 & 0.071 \\ \hline
        AdaBoost & 91 & 1.11E-16 & 0.080 \\ \hline
        RBF SVC & 87.2 & N/A & 1.043 \\ \hline
        GaussianNB & 85.4 & 1.11E-16 & 0.009 \\ \hline
        DecisionTree & 88.4 & 1.11E-16 & 0.007 \\ \hline
    \end{tabular}
\end{table}

\begin{table}[!b]
    \centering
    \caption{Left-right task trained on large grid and tested on pro-antisaccade}
    \label{tab:LRLGPA}
    \begin{tabular}{|l|l|l|l|l|}
    \hline
        Model & Accuracy (\%) & Standard Deviation & Mean Run time (s) \\ \hline
        XGBoost & 96.5 & 1.11E-16 & 0.050  \\ \hline
        GradientBoost & 96.0 & 6.07E-4 & 0.0123 \\ \hline
        AdaBoost & 95.0 & N/A & 0.138 \\ \hline
        RandomForest & 95.0 & 3.57E-4 & 0.085 \\ \hline
        DecisionTree & 91.9 & N/A & 0.006 \\ \hline
        LinearSVC & 90.8 & 1.66E-4 & 0.006 \\ \hline
        RBF SVC & 86.3 & 1.11E-16 & 1.892  \\ \hline
        GaussianNB & 83.7 & N/A & 0.013  \\ \hline
    \end{tabular}
\end{table}

\subsection{Comparison of PA-PA, LG-LG, PA-LG and LG-PA}

\subsubsection{PA-PA vs LG-LG}

Looking at Tables \ref{tab:LRA} and \ref{tab:LRLGP}, we can see that in all 8 models, the prediction accuracies for models trained and tested on pro-antisaccade data were greater than those trained and tested on large grid data. The average difference in accuracy, as shown in Table \ref{tab:data}, was around 1.8\%. Despite this difference, the accuracies of models trained and tested on large grid remained relatively high; all the tested models except GaussianNB and RBG SVC, have accuracies that are higher than or equal to 92\%. On another note, data in Table \ref{tab:LRA} resembles results found by \cite{kastrati2021eegeyenet}, proving the reproducibility of their work.

\subsubsection{PA-LG vs LG-PA}

Comparing Tables \ref{tab:LRALGP} and \ref{tab:LRLGPA}, we see that except for RBF SVC and GaussianNB, models trained on large grid data have higher accuracies when tested on pro-antisaccade data compared to the reverse. As shown in Table \ref{tab:data}, the average improvement in accuracy is around 1.9\%. This result aligns with the hypothesis that in general, models trained on finer-grain data will perform better when covariate shifts are applied, specifically in terms of varying data complexity. It also aligns with previous work regarding the influence of higher complexity EEG data on machine learning \cite{burns2015combining}.

\subsubsection{Implications of LG-PA $>$ PA-LG in General}

Although previous eye-tracking classifiers report higher accuracies ($95.5\%\pm4.6\%$), their models assume distributionally similar data \cite{nilsson2016screening}. The models trained on large grid in this study, on the other hand, retain accuracy even after a distributional shift. This means that regardless of distributional patterns, large grid trained models are able to identify similarity traits in eye-tracking data. Therefore, in general, models trained on vector-based data, rather than binary-classified data, are more adaptable to day-to-day EEG-ET data collection. 

\begin{table} [!b]

    \centering
    \caption{Accuracy of machine learning models on left-right classification. PA-PA indicates trained on pro-antisaccade and tested on pro-antisaccade, LG-LG indicates trained on large grid, tested on large grid, PA-LG indicates trained on pro-antisaccade, tested on large grid, LG-PA indicates trained on large grid, tested on pro-antisaccade.}
    \label{tab:data}
    
    \begin{tabular}{|
        >{\columncolor[HTML]{FFDFA1}}l |l|l|l|l|}
        \hline
        \cellcolor[HTML]{6BACAF}Models       &
        \cellcolor[HTML]{6BACAF}PA-PA (\%) &
        \cellcolor[HTML]{6BACAF}LG-LG (\%) &
        \cellcolor[HTML]{6BACAF}PA-LG (\%) &  \cellcolor[HTML]{6BACAF}LG-PA (\%) \\ \hline
        {\color[HTML]{000000} \textbf{XGBoost}}       & 97.9 & 95.5 & 93.4 & 96.5 \\ \hline
        {\color[HTML]{000000} \textbf{GradientBoost}} & 97.4 & 95.6 & 92.1 & 96.0 \\ \hline
        {\color[HTML]{000000} RandomForest}  & 96.5 & 94.6 & 91.2 & 95.0 \\ \hline
        {\color[HTML]{000000} AdaBoost}      & 96.3 & 93.5 & 91.0 & 95.0 \\ \hline
        {\color[HTML]{000000} DecisionTree}  & 96.2 & 92.0 & 88.4 & 91.9 \\ \hline
        {\color[HTML]{000000} \textbf{LinearSVC}}    & 92.0 & 92.0 & 91.4 & 90.8 \\ \hline
        {\color[HTML]{000000} RBF SVC}       & 89.4 & 88.9 & 87.2 & 86.3 \\ \hline
        {\color[HTML]{000000} GaussianNB}    & 87.7 & 87.2 & 85.4 & 83.7 \\ \hline
        {\color[HTML]{000000} Average}      & 94.2 & 92.4 & 90.0 & 91.9 \\ \hline
    \end{tabular}
\end{table}

\section{Discussion}
This paper utilizes the benchmark data from the EEGEyeNet dataset and 8 machine learning models to create left-right classifiers trained on both coarse-grain and fine-grain data. The accuracies of the models are then tested and compared using two different testing sets of different distributional complexity. In this way, the effects of a covariate distributional shift on the performance of the machine learning models are observed. This provides useful insights into the experimental models and the nature of the data. 

\subsection{Models Trained on Pro-Antisaccade}

As expected from the results of \cite{kastrati2021eegeyenet}, making left-right predictions using data from the pro-antisaccade paradigm is highly accurate. This is shown by Table \ref{tab:data} as the average accuracy of the models trained using pro-antisaccade data are 94.2\% and 90\%. Although these models are accurate, there is a lack of consistency because the average dropped 4.2\% after covariate shifting. This means that models trained on pro-antisaccade data are susceptible to distributional shifts. All of the classifiers dropped in accuracy. DecisionTree, RandomForest, XGBoost GradientBoost, and AdaBoost were the most susceptible, dropping in accuracy by 7.8\%, 5.3\%, 4.5\% 5.3\%, and 5.3\% respectively. RBF SVC, GuassianNB, and LinearSVC were the least susceptible, dropping in accuracy by 2.2\%, 2.3\%, and 0.6\% respectively.

\subsection{Models Trained on Large Grid}

As shown in the results, making left-right classification predictions using data from the large grid paradigm has similarly high accuracies as the predictions made by models trained on pro-antisaccade data. The performance of these machine learning models is, however, more consistent as shown by comparatively minimal decrease in accuracy after covariate shifting. While the acccuracy of models trained on pro-antisaccade data dropped by 4.2\% on average after covariate shifting, classifiers trained using large grid data dropped in accuracy only by 0.5\%. In fact, some of the classifiers increased in accuracy when tested on data from a different, simpler distributional pattern. DecisionTree, RBF SVC, GaussianNB, and LinearSVC decreased in accuracy by 0.1\%, 2.6\%, 3.5\%, and 1.2\% respectively. RandomForest, XGBoost, GradientBoost, and AdaBoost increased in accuracy by 0.4\%, 1\%, 0.4\%, and 1.5\% respectively. 

\subsection{Comparison of the Models and the Best Model for EEG-ET Classification}
 
DecisionTree was the most susceptible classifier when trained on coarse-grain data and did not really change in accuracy when trained on fine-grain data. RandomForest, XGBoost, GradientBoost, and AdaBoost were all high susceptible machine learning models when trained on course-grain data. They were all, however, the best performing when trained on fine-grain data. This suggests that data distribution and representation is most important for DecisionTree, RandomForest, XGBoost, GradientBoost, and AdaBoost. Out of these 5 classifiers XGBoost and GradientBoost performed the best, having high accuracies as well as performing the best after covariate shifting. 
 
As shown in Table \ref{tab:data}, RBF SVC, GaussianNB, and LinearSVC dropped in accuracy a somewhat consistent amount between PA-LG and LG-PA. This shows that regardless of coarse-grain and fine-grain data, these classifiers will drop in accuracy after a covariate shift is applied. Out of these classifiers, LinearSVC was the least susceptible to distributional shifts.

To summarize the results, the most consistent classifier, regardless of coarse-grain and fine-grain data, is LinearSVC whereas the best performing classifiers, after covariate distributional shifting, are XGBoost and GradientBoost, but specifically when trained on fine-grain data. Since the benchmark results of this classification task were already relatively accurate, the variation in performance was not very large. For a benchmark with less accurate results, the variation can only increase, magnifying the results found in this study. These results portray the importance of fine- versus coarse-grain data as well as the types of machine learning models that should be used for EEG-ET classification given a dataset of some consistent/random distributional complexity. We encourage future EEG and eye-tracking research to keep fine- versus coarse-grain data in mind as well as in general, be conducted utilizing fine-grain data collection methods to more accurately emulate real-world stimuli.

\subsection{Future Recommendations and Improvements}
To further advance the work provided in this study, three steps are highlighted for future exploration. 

Although deep learning models were excluded due to inconsistent results \cite{lotte2018review}, the main reason was due to restraints in time and resources. As deep learning models, especially CNN, have shown promising results in regards to EEG-ET classification, it is important to thoroughly explore the effects of fine- versus coarse-grain data on the robustness of such models. 

Additionally, the angle value is currently the only indicator used to determine whether the saccade is towards the left or the right. The amplitude was completely ignored in data processing. The amplitude could be incorporated by using it as a weighting/scaling factor to indicate the left-right extension and solve this as a regression problem. Based on the predictions made by the regression value, we can then classify it as either left or right utilizing finer-grain data.

Furthermore, a testing set was not properly compiled as a dataset made up of both fine-grain and coarse-grain data, in other words a dataset of purely unfamiliar data, was not used. An idea to make this paper more comprehensive would be to pool both complex and simple data into one dataset and then train/test the machine learning models on that. This would confirm whether it is better to train a machine learning model using coarse-grain, binary-classified data or fine-grain, vector-based data to classify EEG-ET data after a covariate distributional shift is applied. 

It is also worth mentioning that our results are derived from training only 8 machine learning models for the specific application of EEG-ET gaze classification; thus, further investigation may be needed to confirm whether the trend will persist for other applications of machine and deep learning models.

\section{Conclusion}
The motivation behind this work was to check whether training machine learning models on finer-grain data leads to more robust models. Our results indicated that when tested on data of similar data-collection complexity, prediction accuracy does not increase. On the other hand, models trained on the fine-grain, vector-based eye-tracking data performed better in general than those trained on the coarse-grain, binary-classified data after a covariate shift is applied. This suggests that training models on vector-based, fine-grain data is more reliable for building practical, day-to-day Human-Computer Interfaces as opposed to binary-classified, coarse-grain data.

\newpage

\section{Appendix}

\subsection{Appendix A}

After emailing Ard Kastrati, Ard verified that for angle $\alpha$ in radians:

$\alpha = 0$ is right

$\alpha = \frac{\pi}{2}$ is down

$\alpha = \pi$ is left

$\alpha = -\frac{\pi}{2}$ is up
\\

\subsection{Appendix B}

The models were trained and tested on this environment settings:

OS: Mac 12.2.1

Cuda: 9.0, Cudnn: v7.03

Python: 3.9.0

cleverhans: 2.1.0

Keras: 2.2.4

tensorflow-gpu: 1.9.0

numpy: 1.22.1

keras: 2.2.4

scikit-learn  1.0.2

scipy 1.8.0

The total space occupied by the dataset on the device is 69.0574 GB, and the total time for training and testing was 30 mins on average.


\begin{thebibliography}{8}

\bibitem{bashivan2016mental}
    Bashivan, P., Rish, I., Heisig, S. (2016). Mental state recognition via wearable EEG. arXiv preprint arXiv:1602.00985.

\bibitem{lotte2015signal}
    Lotte, F. (2015). Signal processing approaches to minimize or suppress calibration time in oscillatory activity-based brain–computer interfaces. Proceedings of the IEEE, 103(6), 871-890.

\bibitem{foster2017alpha}
    Foster, J. J., Sutterer, D. W., Serences, J. T., Vogel, E. K., Awh, E. (2017). Alpha-band oscillations enable spatially and temporally resolved tracking of covert spatial attention. Psychological science, 28(7), 929-941.
    
\bibitem{pedregosa2011scikit}
    Pedregosa, F., Varoquaux, G., Gramfort, A., Michel, V., Thirion, B., Grisel, O., Blondel, M., Prettenhofer, P., Weiss, R., Dubourg, V., Vanderplas, J., 2011. Scikit-learn: Machine learning in Python. the Journal of machine Learning research, 12, pp.2825-2830.

\bibitem{jadhav2016comparative}
    Jadhav, S. D., Channe, H. P. (2016). Comparative study of K-NN, naive Bayes and decision tree classification techniques. International Journal of Science and Research (IJSR), 5(1), 1842-1845.

\bibitem{duda1973pattern}
    Duda, R. O., Hart, P. E. (1973). Pattern classification and scene analysis (Vol. 3, pp. 731-739). New York: Wiley.

\bibitem{higginson2000ecological}
    Higginson, C. I., Arnett, P. A., Voss, W. D. (2000). The ecological validity of clinical tests of memory and attention in multiple sclerosis. Archives of Clinical Neuropsychology, 15(3), 185-204.

\bibitem{marcotte2010neuropsychology}
    Marcotte, T. D., Scott, J. C., Kamat, R., Heaton, R. K. (2010). Neuropsychology and the prediction of everyday functioning. The Guilford Press.

\bibitem{wilson1993}
    Wilson, B. A. (1993). Ecological validity of neuropsychological assessment: Do neuropsychological indexes predict performance in everyday activities?. Applied and Preventive Psychology, 2(4), 209-215.

\bibitem{kastrati2021eegeyenet}
    Kastrati, A., Płomecka, M. M. B., Pascual, D., Wolf, L., Gillioz, V., Wattenhofer, R., Langer, N. (2021). EEGEyeNet: a Simultaneous Electroencephalography and Eye-tracking Dataset and Benchmark for Eye Movement Prediction. arXiv preprint arXiv:2111.05100.

\bibitem{pfeiffer2020eye}
    Pfeiffer, J., Pfeiffer, T., Meißner, M., Weiß, E. (2020). Eye-tracking-based classification of information search behavior using machine learning: evidence from experiments in physical shops and virtual reality shopping environments. Information Systems Research, 31(3), 675-691.

\bibitem{thapaliya2018evaluating}
    Thapaliya, S., Jayarathna, S., Jaime, M. (2018, December). Evaluating the EEG and eye movements for autism spectrum disorder. In 2018 IEEE international conference on big data (Big Data) (pp. 2328-2336). IEEE.

\bibitem{sotoodeh2021preserved}
    Sotoodeh, M. S., Taheri‐Torbati, H., Hadjikhani, N., Lassalle, A. (2021). Preserved action recognition in children with autism spectrum disorders: Evidence from an EEG and eye‐tracking study. Psychophysiology, 58(3), e13740.

\bibitem{kang2020identification}
    Kang, J., Han, X., Song, J., Niu, Z., Li, X. (2020). The identification of children with autism spectrum disorder by SVM approach on EEG and eye-tracking data. Computers in biology and medicine, 120, 103722.
    
\bibitem{qian2021two}
    Qian, P., Zhao, Z., Chen, C., Zeng, Z.,  Li, X. (2021, November). Two Eyes Are Better Than One: Exploiting Binocular Correlation for Diabetic Retinopathy Severity Grading. In 2021 43rd Annual International Conference of the IEEE Engineering in Medicine and Biology Society (EMBC) (pp. 2115-2118). IEEE.

\bibitem{wu2022learning}
    Wu, F., Mai, W., Tang, Y., Liu, Q., Chen, J., Guo, Z. (2022). Learning Spatial-Spectral-Temporal EEG Representations with Deep Attentive-Recurrent-Convolutional Neural Networks for Pain Intensity Assessment. Neuroscience, 481, 144-155.

\bibitem{qu2018personalized}
    Qu, X., Hall, M., Sun, Y., Sekuler, R.,  Hickey, T. J. (2018). A Personalized Reading Coach using Wearable EEG Sensors-A Pilot Study of Brainwave Learning Analytics. In CSEDU (2) (pp. 501-507).
    
\bibitem{freund1997decision}
    Freund, Y., Schapire, R. E. (1997). A decision-theoretic generalization of on-line learning and an application to boosting. Journal of computer and system sciences, 55(1), 119-139.

\bibitem{bhuvaneswari2013support}
    Bhuvaneswari, P., Kumar, J. S. (2013). Support vector machine technique for EEG signals. International Journal of Computer Applications, 63(13).

\bibitem{carrion2021analysis}
    Carrión-Ojeda, D., Fonseca-Delgado, R., Pineda, I. (2021). Analysis of factors that influence the performance of biometric systems based on EEG signals. Expert Systems with Applications, 165, 113967.

\bibitem{aydemir2014decision}
  Aydemir, O., Kayikcioglu, T. (2014). Decision tree structure based classification of EEG signals recorded during two dimensional cursor movement imagery. Journal of neuroscience methods, 229, 68-75.

\bibitem{qian2020multi}
    Qian, P., Zhao, Z., Liu, H., Wang, Y., Peng, Y., Hu, S., ...  Zeng, Z. (2020, July). Multi-target deep learning for algal detection and classification. In 2020 42nd Annual International Conference of the IEEE Engineering in Medicine and Biology Society (EMBC) (pp. 1954-1957). IEEE.

\bibitem{qu2018eeg}
    Qu, X., Sun, Y., Sekuler, R.,  Hickey, T. (2018, October). EEG markers of STEM learning. In 2018 IEEE Frontiers in Education Conference (FIE) (pp. 1-9). IEEE.

\bibitem{edla2018classification}
    Edla, D. R., Mangalorekar, K., Dhavalikar, G., Dodia, S. (2018). Classification of EEG data for human mental state analysis using Random Forest Classifier. Procedia computer science, 132, 1523-1532.

\bibitem{tiwari2019multiclass}
    Tiwari, A., Chaturvedi, A. (2019, November). A multiclass EEG signal classification model using spatial feature extraction and XGBoost algorithm. In 2019 IEEE/RSJ International Conference on Intelligent Robots and Systems (IROS) (pp. 4169-4175). IEEE.

\bibitem{satapathy2017eeg}
    Satapathy, S. K., Dehuri, S., Jagadev, A. K. (2017). EEG signal classification using PSO trained RBF neural network for epilepsy identification. Informatics in Medicine Unlocked, 6, 1-11.

\bibitem{kumar2012analysis}
    Kumar, J. S., Bhuvaneswari, P. (2012). Analysis of Electroencephalography (EEG) signals and its categorization–a study. Procedia engineering, 38, 2525-2536.

\bibitem{klaib2021eye}
    Klaib, A. F., Alsrehin, N. O., Melhem, W. Y., Bashtawi, H. O., Magableh, A. A. (2021). Eye tracking algorithms, techniques, tools, and applications with an emphasis on machine learning and Internet of Things technologies. Expert Systems with Applications, 166, 114037.

\bibitem{qu2020identifying}
    Qu, X., Liukasemsarn, S., Tu, J., Higgins, A., Hickey, T. J.,  Hall, M. H. (2020). Identifying clinically and functionally distinct groups among healthy controls and first episode psychosis patients by clustering on EEG patterns. Frontiers in psychiatry, 938.

\bibitem{plancher2012using}
    Plancher, G., Tirard, A., Gyselinck, V., Nicolas, S., Piolino, P. (2012). Using virtual reality to characterize episodic memory profiles in amnestic mild cognitive impairment and Alzheimer's disease: influence of active and passive encoding. Neuropsychologia, 50(5), 592-602.

\bibitem{gu2020multi}
    Gu, J., Zhao, Z., Zeng, Z., Wang, Y., Qiu, Z., Veeravalli, B., ...  Chow, P. K. (2020, July). Multi-phase cross-modal learning for noninvasive gene mutation prediction in hepatocellular carcinoma. In 2020 42nd Annual International Conference of the IEEE Engineering in Medicine  Biology Society (EMBC) (pp. 5814-5817). IEEE.

\bibitem{li2006data}
    Li, L., Abu-Mostafa, Y. S. (2006). Data complexity in machine learning.

\bibitem{burns2015combining}
    Burns, T., Rajan, R. (2015). Combining complexity measures of EEG data: multiplying measures reveal previously hidden information. F1000Research, 4.

\bibitem{xu2020multi}
    Xu, K., Zhao, Z., Gu, J., Zeng, Z., Ying, C. W., Choon, L. K., ...  Chow, P. K. (2020, July). Multi-instance multi-label learning for gene mutation prediction in hepatocellular carcinoma. In 2020 42nd Annual International Conference of the IEEE Engineering in Medicine and Biology Society (EMBC) (pp. 6095-6098). IEEE.
    
\bibitem{nilsson2016screening}
    Nilsson Benfatto, M., Öqvist Seimyr, G., Ygge, J., Pansell, T., Rydberg, A., Jacobson, C. (2016). Screening for dyslexia using eye tracking during reading. PloS one, 11(12), e0165508.

\bibitem{lobo2016cognitive}
    Lobo, J. L., Ser, J. D., De Simone, F., Presta, R., Collina, S., Moravek, Z. (2016, September). Cognitive workload classification using eye-tracking and EEG data. In Proceedings of the International Conference on Human-Computer Interaction in Aerospace (pp. 1-8).

\bibitem{sabanci2015classification}
    Sabancı, K., Köklü, M. (2015). The classification of eye state by using kNN and MLP classification models according to the EEG signals.

\bibitem{hollenstein2018zuco}
    Hollenstein, N., Rotsztejn, J., Troendle, M., Pedroni, A., Zhang, C., Langer, N. (2018). ZuCo, a simultaneous EEG and eye-tracking resource for natural sentence reading. Scientific data, 5(1), 1-13.

\bibitem{plochl2012combining}
    Plöchl, M., Ossandón, J. P., König, P. (2012). Combining EEG and eye tracking: identification, characterization, and correction of eye movement artifacts in electroencephalographic data. Frontiers in human neuroscience, 6, 278.

\bibitem{oikonomou2020machine}
    Oikonomou, V. P., Nikolopoulos, S., Kompatsiaris, I. (2020). Machine-learning techniques for EEG data. Signal Processing to Drive Human-Computer Interaction: EEG and eye-controlled interfaces, 145.

\bibitem{roy2019machine}
    Roy, S. (2019). Machine Learning for removing EEG artifacts: Setting the benchmark. arXiv preprint arXiv:1903.07825.

\bibitem{zhang2021eegdenoisenet}
    Zhang, H., Zhao, M., Wei, C., Mantini, D., Li, Z., Liu, Q. (2021). Eegdenoisenet: A benchmark dataset for deep learning solutions of eeg denoising. Journal of Neural Engineering, 18(5), 056057.

\bibitem{langer2017resource}
    Langer, N., Ho, E.J., Alexander, L.M., Xu, H.Y., Jozanovic, R.K., Henin, S., Petroni, A., Cohen, S., Marcelle, E.T., Parra, L.C., Milham, M.P., 2017. A resource for assessing information processing in the developing brain using EEG and eye tracking. Scientific data, 4(1), pp.1-20.

\bibitem{lotte2018review}
    Lotte, F., Bougrain, L., Cichocki, A., Clerc, M., Congedo, M., Rakotomamonjy, A., Yger, F. (2018). A review of classification algorithms for EEG-based brain–computer interfaces: a 10 year update. Journal of neural engineering, 15(3), 031005.

\bibitem{qu2020multi}
    Qu, X., Liu, P., Li, Z., Hickey, T. (2020, October). Multi-class Time Continuity Voting for EEG Classification. In International Conference on Brain Function Assessment in Learning (pp. 24-33). Springer, Cham.
    
\bibitem{qu2020using}
    Qu, X., Mei, Q., Liu, P.,  Hickey, T. (2020, October). Using EEG to distinguish between writing and typing for the same cognitive task. In International Conference on Brain Function Assessment in Learning (pp. 66-74). Springer, Cham.

\bibitem{roy2019deep}
    Roy, Y., Banville, H., Albuquerque, I., Gramfort, A., Falk, T. H., Faubert, J. (2019). Deep learning-based electroencephalography analysis: a systematic review. Journal of neural engineering, 16(5), 051001.

\bibitem{craik2019deep}
    Craik, A., He, Y.,  Contreras-Vidal, J. L. (2019). Deep learning for electroencephalogram (EEG) classification tasks: a review. Journal of neural engineering, 16(3), 031001.
    
\end{thebibliography}
\end{document}